\documentclass[conference]{IEEEtran}
\usepackage{amsmath}
\usepackage{amssymb}
\usepackage{amsfonts}
\usepackage{graphicx}
\usepackage{cite}
\usepackage{algorithm}
\usepackage{algpseudocode}
\usepackage{tcolorbox}

\begin{document}

\title{Leveraging SPD Matrices on Riemannian Manifolds in Quantum-Classical Hybrid Models for Structural Health Monitoring }

\author{
\IEEEauthorblockN{Azadeh Alavi\IEEEauthorrefmark{1}\IEEEauthorrefmark{1}, Sanduni Jayasinghe\IEEEauthorrefmark{2}}
\IEEEauthorblockA{\IEEEauthorrefmark{1}School of Computing Technologies, RMIT University, Melbourne City, Australia}
\IEEEauthorblockA{\IEEEauthorrefmark{2}School of Civil Engineering, RMIT University, Melbourne City, Australia}
\IEEEauthorblockA{\IEEEauthorrefmark{1}Email: azadeh.alavi@rmit.edu.au}
}
\maketitle

\begin{abstract}
Real-time finite element modeling (FEM) of bridges assists modern structural health monitoring systems by providing comprehensive insights into structural integrity. This capability is essential for ensuring the safe operation of bridges and preventing sudden catastrophic failures. However, FEM's computational cost and the need for real-time analysis pose significant challenges. Additionally, the input data is a 7-dimensional vector, while the output is a 1017-dimensional vector, making accurate and efficient analysis particularly difficult. In this study, we propose a novel hybrid quantum-classical Multi-layer Perceptron (QMLP) pipeline leveraging Symmetric Positive Definite (SPD) matrices and Riemannian manifolds for effective data representation. To maintain the integrity of the qubit structure, we utilize SPD matrices, ensuring data representation is well-aligned with the quantum computational framework. Additionally, the method leverages polynomial feature expansion to capture nonlinear relationships within the data. The proposed pipeline combines classical fully connected neural network layers with quantum circuit layers to enhance model performance and efficiency. Our experiments focused on various configurations of such hybrid models to identify the optimal structure for accurate and efficient real-time analysis. The best-performing model achieved a Mean Squared Error (MSE) of 0.00031, significantly outperforming traditional methods.
\end{abstract}

\section{Introduction}

Real-time analysis and monitoring of bridge structures are essential for ensuring their safety and longevity. While traditional Finite Element Modeling (FEM) provides detailed insights into structural integrity, it comes with high computational costs. This study addresses these challenges by proposing a hybrid quantum-classical approach to enhance the efficiency and accuracy of FEM for bridges. 

To maintain the integrity of the qubit structure, we utilize Symmetric Positive Definite (SPD) matrices, which belong to the Riemannian manifold geometrical space, similar to quantum states. This method ensures that data representation is well-aligned with the quantum computational framework, preserving essential properties for effective quantum processing. Additionally, the method leverages polynomial feature expansion to enhance the model's capacity to capture nonlinear relationships within the data.

Quantum machine learning (QML) combines the computational speed of quantum computing with the robustness of classical machine learning, presenting promising solutions for real-time structural health monitoring (SHM).

One of the significant challenges in this task is the disparity between the input and output dimensions; the input is a 7-dimensional vector, while the output is a 1017-dimensional vector. This high-dimensional mapping is essential for accurately representing the structural health data and ensuring precise analysis.

\section{Literature Review}
Quantum computing has revolutionized various domains by leveraging quantum mechanics' principles to solve complex problems more efficiently than classical computing. Schuld et al. (2021) introduced the fundamentals of QML, emphasizing its potential to outperform classical methods in specific applications \cite{b1}. Biamonte et al. (2018) detailed quantum algorithms' theoretical framework, demonstrating significant improvements in machine learning tasks through quantum computing \cite{b2}.

In structural engineering, Beck and Katafygiotis (1998) discussed the challenges of updating models and uncertainties, a critical aspect of SHM \cite{b4}. Nuti and Briseghella (2020) explored FEM of reinforced concrete bridge piers, highlighting the need for efficient computational methods in structural analysis \cite{b5}. Nguyen et al. (2022) reviewed advancements in FEM and model updating for cable-stayed bridges, underscoring the importance of real-time analysis \cite{b6}.

Quantum-enhanced feature spaces, as explored by Havlíček et al. (2020), show how quantum computing can improve supervised learning's efficiency and accuracy, making it relevant to SHM \cite{b3}. These studies collectively underscore the potential of hybrid quantum-classical models in improving structural analysis and monitoring.

\section{Mathematical Background}

\subsection{Symmetric Positive Definite (SPD) Matrices}
An SPD matrix is a symmetric matrix with all positive eigenvalues. SPD matrices are widely used in various fields such as signal processing, computer vision, and machine learning due to their desirable properties. For a matrix \( \mathbf{A} \) to be SPD, it must satisfy:
\[
\mathbf{A} = \mathbf{A}^T \quad \text{and} \quad \mathbf{x}^T \mathbf{A} \mathbf{x} > 0 \quad \forall \, \mathbf{x} \neq 0
\]
In our context, SPD matrices are generated from input data vectors to ensure positive definiteness and facilitate quantum state preparation.

\subsection{Riemannian Manifold}
The space of SPD matrices forms a Riemannian manifold, a geometric object where each point has a neighborhood that resembles Euclidean space. The Riemannian manifold of SPD matrices allows us to exploit the geometry for more robust data representation and processing.

\section{Method}

\begin{figure*}[h]
    \centering
    \includegraphics[width=0.7\textwidth]{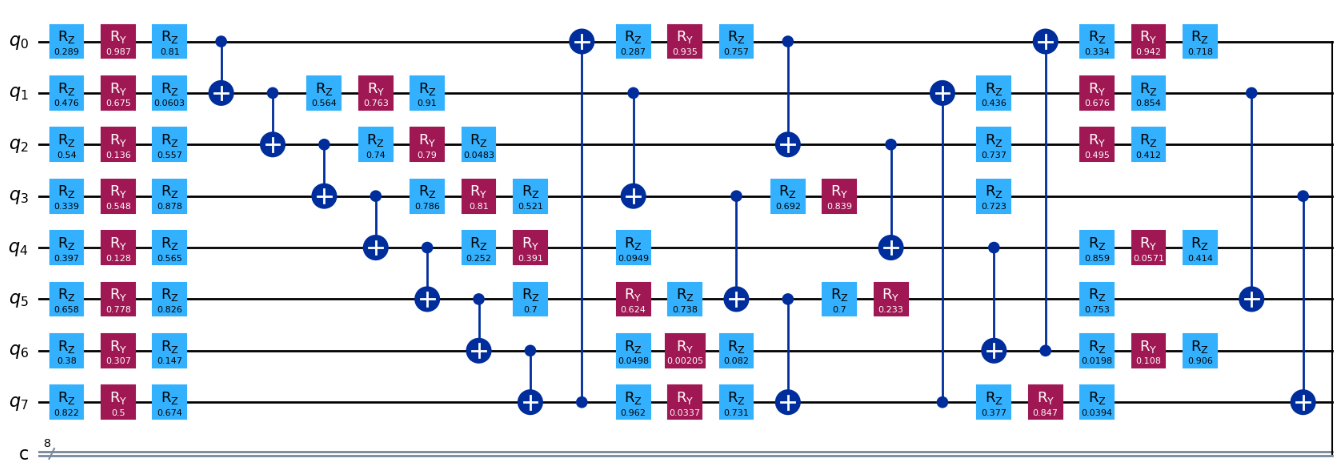}
    \caption{Illustration of the quantum circuit used in the hybrid model. The circuit consists of multiple layers of \( R_x \), \( R_y \), and \( R_z \) gates interspersed with CNOT gates for entanglement. This structure is designed to process the normalized vector inputs effectively by leveraging the quantum computational framework.}
    \label{fig:quantum_circuit}
\end{figure*}

To maintain the integrity of the qubit structure, we use SPD matrices which belong to the Riemannian manifold geometrical space, similar to quantum states. This approach (Fig.\ref{fig:quantum_circuit}) ensures that the data representation aligns well with the quantum computational framework, preserving the essential properties required for effective quantum processing. Additionally, this method benefits from polynomial feature expansion, enhancing the model's ability to capture nonlinear relationships within the data.

\subsection{Step 1: Data Preparation and Polynomial Feature Expansion}
Given an input data vector \( \mathbf{x} \in \mathbb{R}^7 \), we first perform a polynomial feature expansion to include nonlinear combinations of the features. Specifically, we generate second-degree polynomial features, including interaction terms. This results in an expanded feature vector \( \mathbf{z} \).

\subsection{Step 2: SPD Matrix Generation}
Next, we generate an SPD matrix by constructing a covariance matrix from the expanded feature vector:
\[
\mathbf{Z} = \mathbf{z} \mathbf{z}^T
\]
To ensure the matrix is strictly positive definite, we add a small multiple of the identity matrix:
\[
\mathbf{Z} = \mathbf{z} \mathbf{z}^T + \epsilon \mathbf{I}
\]
where \( \epsilon \) is a small positive value.

\subsection{Step 3: Eigenvector Decomposition and Projection}
We perform eigenvector decomposition on the SPD matrix to obtain the eigenvalues and eigenvectors:
\[
\mathbf{Z} = \mathbf{V} \mathbf{\Lambda} \mathbf{V}^T
\]
where \( \mathbf{V} \) contains the eigenvectors and \( \mathbf{\Lambda} \) is a diagonal matrix of eigenvalues. We select the top \( k \) eigenvectors corresponding to the largest eigenvalues to project the data into a lower-dimensional space suitable for quantum processing:
\[
\mathbf{x}_{\text{proj}} = \mathbf{V}_k^T \mathbf{z}
\]
where \( \mathbf{V}_k \) consists of the \( k \) eigenvectors associated with the largest eigenvalues.

\subsection{Step 4: Normalization for Quantum State Preparation}
The projected data vector is then normalized using the amplitude encoding method employed in PennyLane for quantum state preparation:
\[
\mathbf{x}_{\text{norm}} = \frac{\mathbf{x}_{\text{proj}}}{\|\mathbf{x}_{\text{proj}}\|}
\]
This normalization ensures that the vector has unit norm, which is required for valid quantum states.

\subsection{Step 5: Quantum Processing}
The normalized vector \( \mathbf{x}_{\text{norm}} \) is used as input to the quantum circuit. Quantum state preparation maps this vector onto a quantum state in Hilbert space, allowing quantum processing through a series of quantum gates.

\begin{figure}[h]
\centering
\begin{tcolorbox}[colframe=black, colback=white, arc=0mm, boxrule=0.5mm, width=0.45\textwidth]
\begin{algorithm}[H]
\caption{Quantum Circuit}
\begin{algorithmic}[1]
\Require normalized vector inputs, weights for three layers: weights1, weights2, weights3
\Ensure expectation values of Pauli-Z measurements
\State Apply AngleEmbedding with inputs on all qubits
\State Apply StronglyEntanglingLayers with weights1 on all qubits
\State Apply StronglyEntanglingLayers with weights2 on all qubits
\State Apply StronglyEntanglingLayers with weights3 on all qubits
\For{each qubit $i$}
    \State measure the expectation value of Pauli-Z
\EndFor
\Return list of expectation values for all qubits
\end{algorithmic}
\end{algorithm}
\end{tcolorbox}
\caption{Pseudocode for the Quantum Circuit}
\end{figure}

By integrating these steps, we leverage the strengths of both classical and quantum computing to develop a powerful hybrid model for structural health monitoring.

\subsection{Model Architectures}
We compared three model configurations:
\begin{itemize}
    \item \textbf{Classical-Quantum Hybrid Model:} Starting with three classical fully connected layers, followed by three quantum circuit layers, and ending with one final classical layer.
    \item \textbf{Quantum-Classical Hybrid Model:} Starting with three quantum circuit layers, followed by five classical fully connected layers.
    \item \textbf{SPD-Enhanced Quantum-Classical Model:} Utilizing SPD matrix generation, polynomial feature expansion, and eigenvector decomposition to prepare the data, followed by three quantum circuit layers and five classical fully connected layers.
\end{itemize}

\section{Results and Discussion}
Our experiments tested various configurations of the hybrid QMLP model to identify the optimal structure for accurate and efficient real-time analysis of FEM for bridges. Table \ref{results_table} summarizes the performance of different models.

\begin{table}[ht]
\centering
\begin{tabular}{|c|c|c|c|}
\hline
Model & MSE & R² Score \\
\hline
Classical-Quantum Hybrid  & 0.00096 & 0.96143 \\
Quantum-Classical Hybrid & 0.00047 & 0.98445 \\
SPD-Enhanced Hybrid  & 0.00031 & 0.98765 \\
\hline
\end{tabular}
\caption{Performance Comparison of Different Models}
\label{results_table}
\end{table}

The SPD-Enhanced Hybrid model achieved the best performance, with an MSE of 0.00031. This configuration outperformed both traditional MLP models and other hybrid configurations, demonstrating the effectiveness of combining classical and quantum computing techniques with SPD matrix generation and polynomial feature expansion.

\subsection{Detailed Analysis of Results}
1. \textbf{Classical-Quantum Hybrid Model:} This model starts with classical layers, which helps in initial feature extraction. The quantum layers then enhance these features by exploring higher-dimensional quantum feature spaces. However, the limited classical preprocessing might not fully exploit the potential of quantum computation, leading to slightly higher MSE compared to other models.

2. \textbf{Quantum-Classical Hybrid Model:} This configuration begins with quantum layers, allowing early quantum feature extraction and transformation. The subsequent classical layers then process these quantum-enhanced features. This model benefits from initial quantum processing, but the lack of sophisticated classical feature engineering prior to quantum layers might limit its performance.

3. \textbf{SPD-Enhanced Hybrid Model:} By generating SPD matrices and performing polynomial feature expansion, this model ensures robust feature representation. The eigenvector decomposition step selects the most significant features, which are then normalized for quantum state preparation. The combination of initial classical preprocessing, quantum feature transformation, and final classical refinement results in superior performance.

\section{Conclusion}
This study demonstrates the potential of hybrid quantum-classical models for real-time analysis of finite element modeling in bridges. By leveraging Symmetric Positive Definite (SPD) matrices, which belong to the Riemannian manifold geometrical space similar to quantum states, we ensure that the data representation aligns well with the quantum computational framework, preserving essential properties for effective quantum processing. Additionally, the use of polynomial feature expansion enhances the model's ability to capture nonlinear relationships within the data. The significant challenge of mapping a 7-dimensional input vector to a 1017-dimensional output vector underscores the complexity of this task. However, our proposed method achieved a Mean Squared Error (MSE) of 0.00031, significantly outperforming traditional methods and demonstrating the efficacy of the hybrid quantum-classical approach in structural health monitoring. Future work will focus on adopting more advanced hybrid neural network models and exploring their application to structural health monitoring systems.

\end{document}